\documentclass[conference]{IEEEtran}
\usepackage{cite}
\usepackage{url}
\usepackage{orcidlink}
\usepackage{float}
\usepackage{amsmath}
\usepackage{amssymb}
\usepackage{array}
\usepackage{tabularx}
\usepackage{booktabs}
\usepackage{multirow}
\usepackage{colortbl}
\usepackage{graphicx}
\usepackage{xcolor}
\usepackage{soul}
\usepackage{tikz}
\usetikzlibrary{shapes.geometric, arrows.meta, positioning, fit, backgrounds, calc, matrix}
\usepackage{pgfplots}
\usepgfplotslibrary{groupplots}
\pgfplotsset{compat=1.18}

\usepackage[ruled,vlined]{algorithm2e}

\usepackage[figureposition=bottom,tableposition=top]{caption}
\newcommand{\para}[1]{\paragraph{\textnormal{\textbf{#1}.}}}

\begin{document}

\title{Freezing of Gait Prediction Under Spatial Occlusion: An
IMU-Supervised Cross-Modal Distillation Approach}

\author{
\IEEEauthorblockN{Chandan Biswas \orcidlink{0000-0003-4468-7396}}
\IEEEauthorblockA{NeuroAI Fusion Labs\\
Kolkata, India\\
chandan@neuroailabs.in}
\and
\IEEEauthorblockN{Aryan Singh \orcidlink{0009-0003-0495-627X}}
\IEEEauthorblockA{NeuroAI Fusion Labs\\
Kolkata, India\\
aryan@neuroailabs.in}
\and
\IEEEauthorblockN{Anabik Pal
\orcidlink{0000-0001-5737-9056}}
\IEEEauthorblockA{Indian Institute of Science Education and Research\\
Berhampur, Odisha\\
anabikpal@iiserbpr.ac.in}
}

\maketitle

\begin{abstract}
Parkinson's disease is a progressive neurodegenerative disorder in humans characterised by the gradual deterioration of movement control. Automated freezing-of-gait (FOG) detection supports the objective assessment of gait-related motor impairment. Two common approaches are used for the FOG prediction: (i) analysing video recordings of the patient’s movements and (ii) analysing data collected using inertial measurement unit (IMU) wearable sensors attached to the patient’s lower limbs. Video-based approaches may suffer from detection errors during continuous turning-in-place tasks because the lower limbs undergo substantial geometric self-occlusion, which can degrade pose-estimation accuracy. In contrast, IMU-based approaches are generally less affected by visual occlusion; however, they can be difficult to deploy outside clinical or laboratory settings because the sensors must be attached securely and remain in place throughout the assessment.  




Motivated by this, we propose a cross-modal subspace distillation framework to mitigate the limitations of unimodal FOG detection by combining IMU accuracy with video-based practicality. Specifically, we extract invariant latent topologies from a pre-trained kinematic oracle to structurally supervise a non-encoded visual architecture during training. To resolve periods of severe spatial occlusion, a dual-stream visual model probabilistically fuses skeletal graph nodes and continuous spatial pixels, dynamically shifting reliance to uninterrupted pixel boundaries precisely as joint tracking confidence drops. Evaluated against a public, multi-modal sequence dataset of Parkinson's individuals executing continuous $360^\circ$ turns, empirical results demonstrate that applying sensory boundary topologies strictly mitigates tracking evaluation entropy. Consequently, our constrained optimisation confirms that highly precise FOG prediction bounds can be achieved over zero-wearable inference environments.

\end{abstract}


\section{Introduction}
\label{sec:intro}
Parkinson’s disease (PD) is a progressive neurodegenerative disorder that leads to cardinal motor symptoms (bradykinesia, resting tremor, rigidity, postural instability) and frequent gait disturbances such as freezing of gait. 
This disease is the second most common neurodegenerative disease, and the WHO estimated that, in 2019, PD resulted in 5.8 million disability-adjusted life years, an increase of 81\% since 2000, and caused 329,000 deaths, an increase of over 100\% since  2000~\footnote{https://www.who.int/news-room/fact-sheets/detail/parkinson-disease}. Prolonged FOG episodes correlate strongly with a higher probability of falls and injuries. Objective assessment of FOG is an important clinical requirement for evaluating motor symptom progression in PD.

\begin{figure}[t]
    \centering
    \includegraphics[width=0.49\linewidth]{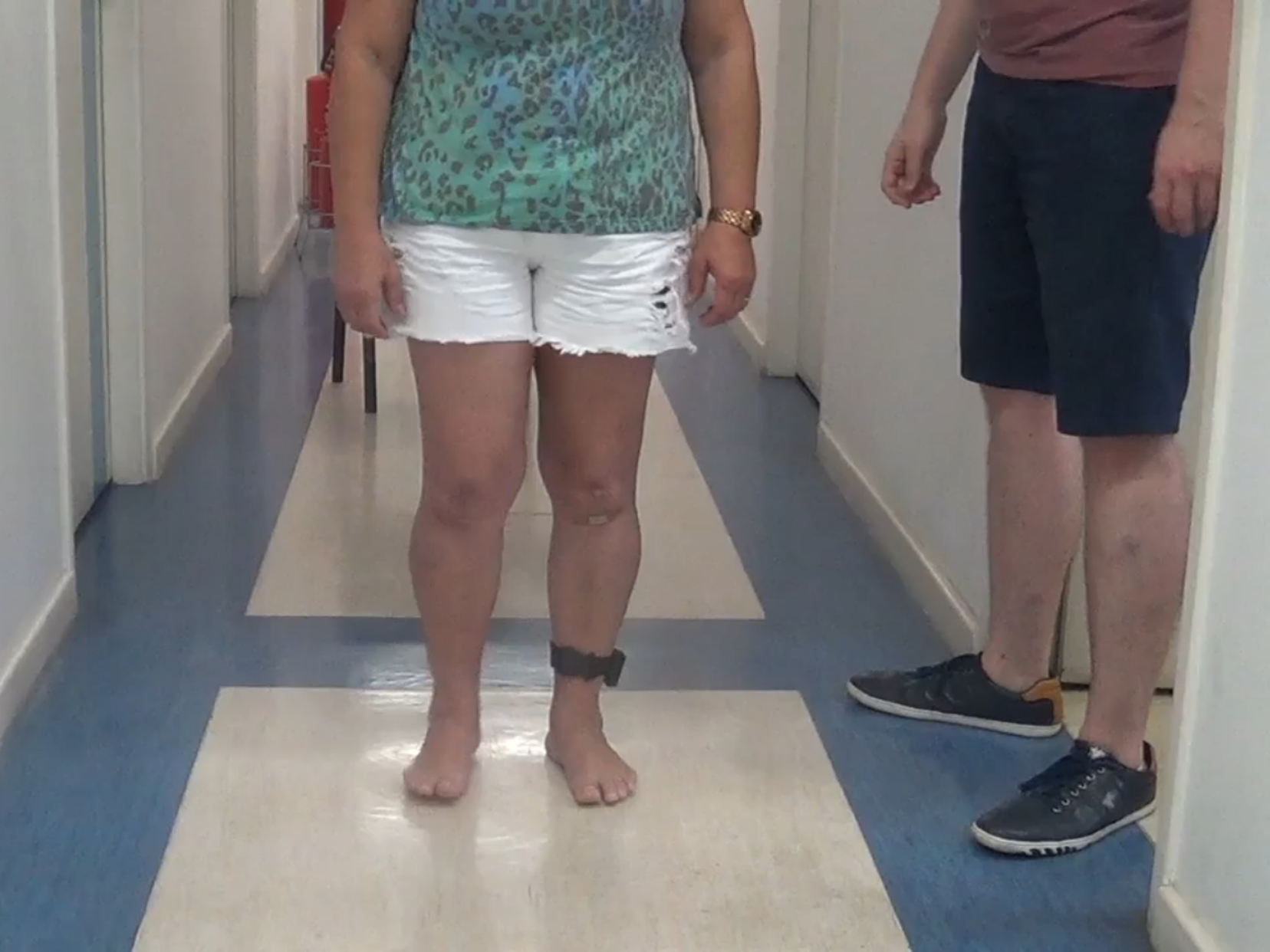}
    \includegraphics[width=0.49\linewidth]{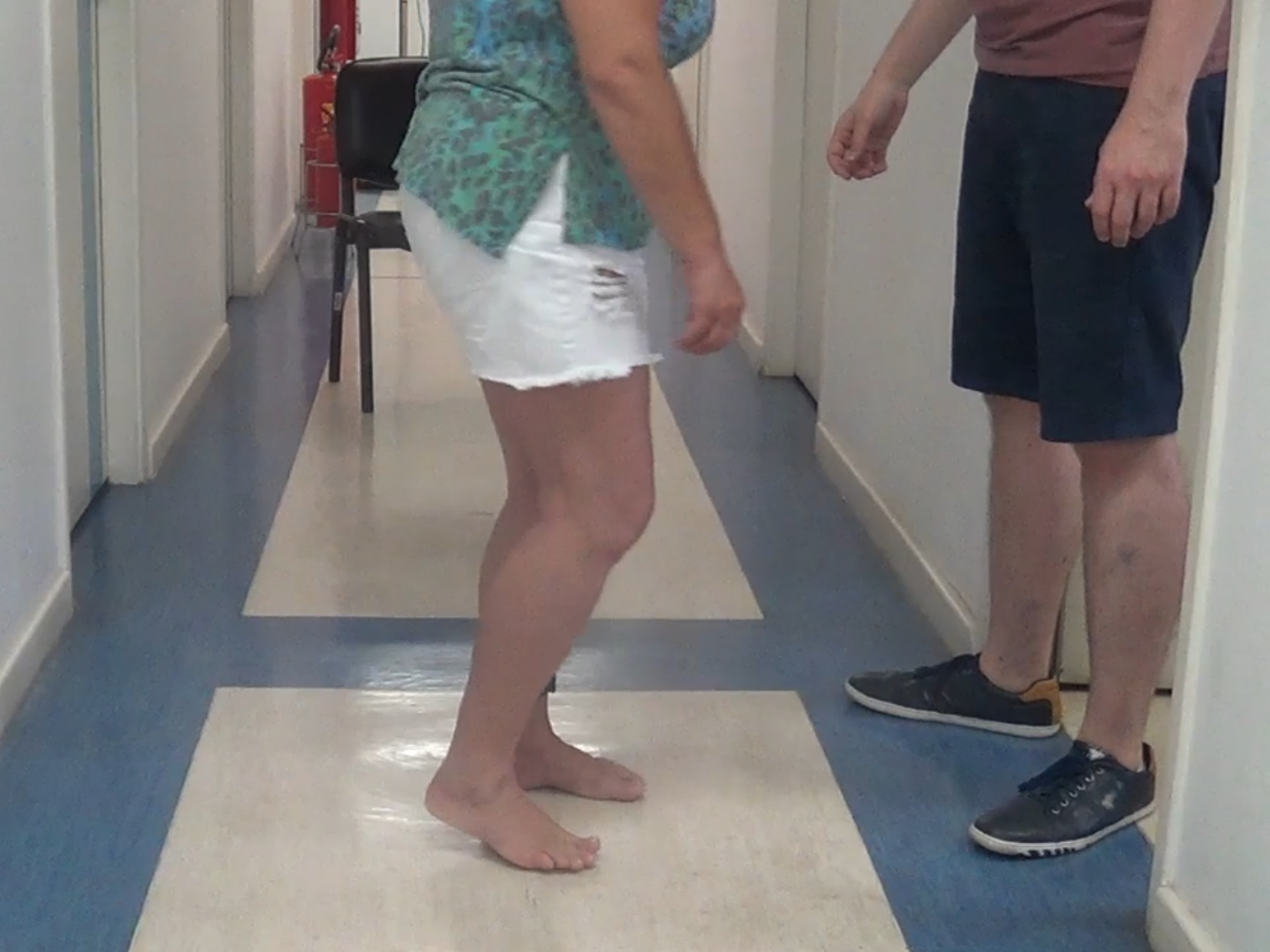}
    \caption{\small{Manifestation of visual self-occlusion during a turning-in-place protocol. }}
    \label{fig:occlusion_demo}
\end{figure}

Recent research shows that technology focused on estimating spatial movement trajectories using physical sensors provides reliable, high-precision FOG assessment results. However, it is challenging to deploy such technology outside controlled laboratory conditions, as the IMU affixed to patients' lower limbs are uncomfortable to wear continuously. In contrast, smartphone-based video acquisition for analysis is a viable option for easy deployment. However, deriving diagnostic kinematic measurements solely from continuous optical fields can introduce substantial errors, particularly during clinical evaluations focused on turning-in-place tasks. As demonstrated in Figure \ref{fig:occlusion_demo}, calculating spatial joint geometry via camera hardware introduces physical uncertainties during periods of spatial self-occlusion. Specifically, when one leg rotates in front of the other, standard skeletal joint algorithms estimate coordinates with diminished confidence limits, resulting in signal noise. A standard learning classifier, completely deprived of deterministic structural points, fails to resolve subtle FOG kinematic characteristics across obscured visual bounds.

To construct an effective defence against these physical ambiguities without assuming an extensive increase in hardware parameters, we formulate an alternative approach exploiting information from distinct data domains. Instead of relying solely on explicit optical patterns to infer motion properties, our workflow utilizes a synchronized multi-modal representation logic during the training phase. We postulate that leveraging a verified hardware signal—such as an IMU embedding subspace containing high task dependency correlations—acts as an explicit mapping mechanism. By projecting a mathematically regularized multi-task constraint, we train an underlying probabilistic video classifier to approximate an expert target metric space rather than modelling individual corrupted geometric constraints sequentially. 
By applying cross-modal latent alignment, the parametrizations acquired on the unified representations constrain visual and tabular classification functions structurally analogous to sensory arrays. Consequently, the diagnostic assessment process performs uninterrupted inference estimates, completely decoupled from explicit tracking boundaries, without imposing direct dependencies on sensor input. 



\noindent \textbf{Contributions:} In summary, the key contributions of this paper are as follows:

\begin{itemize}
    \item We detail a cross-modal subspace alignment training pipeline transferring domain properties from 1-D spatial kinematics directly to unannotated spatio-temporal video subsets. 
    \item We introduce an adaptive structural evaluation limit via parameter-weight fusion distributions based explicitly upon dynamic skeleton coordinate confidences (to solve systematic lower limb tracking corruption conditions probabilistically).
    \item We conduct evaluation experiments isolating metrics resilient to highly sparse clinical variables demonstrating empirical event classification bounds nearing upper bounds measured over attached sensors independent of external observation limitations. 
\end{itemize}
\section{Related Work}
\label{sec:related_work}

This section reviews related work across three research directions: Kinematic and Vision-Based FOG prediction, Latent Space Alignment and Cross-modal Distillation, and Coordinate Representation under Visual Uncertainty. 

\subsection{Kinematic and Vision-Based FOG prediction}
Quantifying rapid temporal movement artifacts, such as the $3-8$ Hz festination signatures characteristic of Freezing of Gait (FOG), typically utilizes hardware sensor networks. Deep learning evaluations of Inertial Measurement Unit (IMU) arrays validate high correlation limits to medical benchmarks \cite{bachlin2010wearable}. Operating directly on physical inertial bounds allows standard classification margins to routinely exceed a predictive fidelity of $90\%$ without subjective filtering \cite{li2020improved}. The public repository formalized by \cite{souza2022public} verifies these operational limits but reiterates the inherent deployment friction regarding continuous device attachment at inference time. 

Transitioning from sensory to markerless video limits restricts objective spatial observation \cite{mancini2021measuring}. Studies examining RGB frame derivations generally constrain visual input to strictly frontal or sagittal movement vectors, minimizing temporal coordinate crossing. Applying analogous evaluation paradigms directly to non-linear sequences—such as a dynamic $360^\circ$ continuous rotational condition—causes systematic observation breakdown within unmodified computer vision classifiers due to an inability to mathematically reconcile heavy geometric occlusion points.

\subsection{Latent Space Alignment and Cross-modal Distillation}
When disparate observational domains generate unequal functional precision, restricting structural entropy in the subordinate domain improves limits generalization. The mechanism of parameter distillation, fundamentally defining limits by transferring distribution outputs across heterogeneous modalities \cite{hinton2015distilling}, constitutes the mathematical baseline of teacher-student topologies. Scaling beyond logit regressions, joint representation environments evaluate probability estimations utilizing symmetrical contrastive distance \cite{radford2021learning}. Applied specifically to action recognition, architectures minimizing vector variations mapping RGB streams into analogous positional domains—such as Audio or skeletal topologies \cite{thoker2021cross}—enable zero-dependency modality inference limits. In standard literature, dual-stream architectures process components conditionally with equal likelihood metrics. We extend this by configuring an inherently rigid IMU embedding as an invariant parameter subset, constraining sequence derivations purely based upon a rigid spatial Oracle alignment formulated directly across dynamic turning vectors. 

\subsection{Coordinate Representation under Visual Uncertainty}
Defining physical topologies for non-pixel networks usually adopts coordinate representations encoded mathematically via Spatio-Temporal Graph Convolutional Networks (ST-GCN) \cite{yan2018spatial}. Baseline ST-GCN parameters assume an uninterrupted series of structural input locations modeling adjacency weights conditionally over fixed node matrices. Such modeling paradigms degrade strictly monotonically when node estimations operate beneath probabilistic noise bounds. Deterministic graph propagation does not possess dynamic bounds resolving variables when target confidence margins estimate null configurations ($C \approx 0$). Work regarding noisy topological extraction normally operates through mathematical imputation bounds, modeling linear node continuation states independent of raw pixel parameters \cite{song2023modeling}. Diverging from rigid topological corrections, our parameter configuration accepts spatial uncertainty physically through expected node activation ranges. Treating spatial topology deterministically across unobstructed bounds and relying proportionately upon continuous raw matrices exclusively when tracking limits fall below minimal prediction margins standardizes objective extraction through complete blockage intervals.
\section{Proposed Method}
\label{sec:proposed}

In this section, we describe the details of our proposed cross-modal learning framework for FOG prediction. The framework is designed to train a vision-and-language model under the strict supervision of an informative kinematic subspace (derived from inertial sensors) to compensate for visual information loss during periods of severe physical occlusion. 

\subsection{Problem Formulation and Synchronisation}
Let $\mathcal{D} = \{ (\mathbf{X}_V^{(i)}, \mathbf{X}_I^{(i)}, \mathbf{x}_T^{(i)}, y^{(i)}) \}_{i=1}^N$ be the synchronised training dataset consisting of $N$ data instances. For each data instance $i$, $\mathbf{X}_V^{(i)} \in \mathbb{R}^{F \times H \times W \times 3}$ represents a sequence of $F$ visual frames in a spatial resolution of $H \times W$. Let $\mathbf{X}_I^{(i)} \in \mathbb{R}^{S \times 6}$ represent the temporally synchronised sequence of $S$ inertial measurements from the triaxial accelerometer and gyroscope. 

To introduce contextual metadata (e.g., age, medication state, clinical rating scale) into the formulation, let $\mathbf{x}_T^{(i)}$ represent an encoded text prompt denoting the patient's clinical profile. The variable $y^{(i)} \in \{0, 1\}$ represents the ground-truth cluster label for the FOG class. During inference on the client side, exclusively the visual sequence $\mathbf{X}_V$ is available. The tabular clinical metadata ($\mathbf{x}_T$) is strictly unavailable during deployment to guarantee fully unobtrusive, zero-dependency evaluations. Consequently, the research objective is to estimate a parameterized transformation of the visual inputs such that they project onto an informative kinematic and clinical latent subspace, optimizing the prediction of $y^{(i)}$ without requiring continuous sensory or tabular inputs at inference.

\subsection{The Kinematic Teacher Oracle}
Due to the physical characteristics of the FOG condition, a dedicated kinematic representation contains a highly discriminative subspace. We leverage an independently pre-trained deep neural network as an expert extractor. Let $E_I(\cdot; \theta_I)$ denote the function projecting the inertial inputs into a $p$-dimensional Euclidean space as follows:
\begin{equation}
    \mathbf{z}_I = E_I(\mathbf{X}_I; \theta_I), \quad \mathbf{z}_I \in \mathbb{R}^p
    \label{eq:imu_extractor}
\end{equation}
Throughout our multi-modal training procedure, we treat the parameters $\theta_I$ as constant (frozen). The objective is to utilise $\mathbf{z}_I$ as the reference vector capturing complete kinematic properties, specifically forcing the latent representations generated by the non-encoded vision and textual networks ($\mathbf{z}_V$
and $\mathbf{z}_T$) to align geometrically with this hardware boundary during optimisation.

\subsection{Probabilistic Dual-Stream Vision Model}
Visual assessment of the turning-in-place task suffers from frequent self-occlusions, particularly when one leg crosses the other. Because a strict deterministic mapping from skeletal coordinates to a label introduces error during periods of heavy occlusion, we frame the extraction of visual features from a probabilistic perspective. 

Let the visual stream be composed of two parallel embeddings. A spatio-temporal Convolutional 3D model ($E_{rgb}$) computes a latent representation over dense pixels yielding $\mathbf{v}_{rgb} \in \mathbb{R}^p$. Simultaneously, a skeleton pose estimator extracts a graph of joint coordinates over $F$ frames. We use a Graph Convolutional Network ($E_{sk}$) to yield $\mathbf{v}_{sk} \in \mathbb{R}^p$. 

However, since pose tracking fails systematically under self-occlusion, treating $\mathbf{v}_{sk}$ equally across all sequences leads to information degradation. We consider the skeletal output to be corrupted by noise, modelled as an uncertainty random variable. 
Let $c_{j}^t \in [0, 1]$ be the observed confidence score from the pose estimation framework for the $j^{\text{th}}$ tracking joint at frame $t$. We hypothesise that the occurrence of valid topological information follows a distribution proportionate to the observed coordinate confidence. We model the estimated validity $C_V$ of a window of skeletal data as a simple expected confidence value:
\begin{equation}
    \mathbb{E}[C_V] = \frac{1}{F \cdot K} \sum_{t=1}^{F} \sum_{j=1}^{K} c_{j}^t 
    \label{eq:pose_expectation}
\end{equation}
where $K$ refers to the number of skeletal tracking nodes. 
We thus treat the actual feature vector for the given data instance as an expected value from a weighted mixture between the dense pixel stream (resistant to missing nodes) and the skeletal stream (susceptible to occlusion) yielding the unified vision embedding $\mathbf{z}_V \in \mathbb{R}^p$, formally defined as:
\begin{equation}
    \mathbf{z}_V = \alpha \cdot \mathbf{v}_{sk} + (1 - \alpha) \cdot \mathbf{v}_{rgb}, 
    \label{eq:vision_fusion}
\end{equation}
where the combination coefficient $\alpha$ is derived from a parameterized sigmoid transformation evaluated over the tracking validity $\mathbb{E}[C_V]$:
\begin{equation}
    \alpha = \frac{1}{1 + \exp(-(\omega_c \cdot \mathbb{E}[C_V] + b_c))},
    \label{eq:alpha_gate}
\end{equation}
where $\omega_c$ and $b_c$ act as explicitly trainable scaling and shift parameters updated alongside $\Theta$. As geometric spatial overlap deprives the skeleton matrices of continuous resolution bounds ($\mathbb{E}[C_V]$ is low), the adaptive scalar threshold evaluates past shifting thresholds causing $\alpha$ to systematically decay towards zero. Consequently, $\mathbf{z}_V$ transitions mapping inference boundaries dynamically into unbroken pixel topologies. 

\subsection{Supervised Cross-Modal Subspace Distillation}
\label{subsec:supervised_distillation}

To address the situation where explicitly annotated temporal bounds of FOG micro-movements are difficult to estimate purely from video, we supervise the fused vision embedding $\mathbf{z}_V$ using the kinematics oracle $\mathbf{z}_I$. In parallel, let $\mathbf{z}_T \in \mathbb{R}^p$ denote a vector obtained from $E_{txt}(\mathbf{x}_T; \theta_T)$, modeling the metadata subspace of the text features. 

Applying standard contrastive alignment unconditionally across random batch segments generates false negative structural penalization, where distinct sequence bounds matching equivalent target pathology (i.e., multiple FOG blocks within the same batch) mathematically repulse. To resolve this, we learn a parameterized joint-space function applying a multi-target Supervised Contrastive Loss (SupCon). 

Let $\mathcal{P}(i) = \{ k \in \{1 \dots M\} \mid y^{(k)} = y^{(i)} \}$ define the explicit index set identifying matching inference classes relative to an evaluated anchor instance $(i)$ spanning a batch of size $M$. Integrating the cardinality $|\mathcal{P}(i)|$, we define the temperature-scaled similarity alignment function targeting the IMU representation subspace independent of inter-instance isolation penalties as:
\begin{equation}
    \mathcal{L}_{sup}^{V \rightarrow I} = \sum_{i=1}^{M} \frac{-1}{|\mathcal{P}(i)|} \sum_{k \in \mathcal{P}(i)} \log \frac{\exp(\mathbf{z}_V^{(i)} \cdot \mathbf{z}_I^{(k)} / \tau)}{\sum_{j=1}^{M} \exp(\mathbf{z}_V^{(i)} \cdot \mathbf{z}_I^{(j)} / \tau)}.
    \label{eq:supcon_vision}
\end{equation}

Similarly, to force the vision manifold to become sensitive to varying states of the clinical profiles of the patient cohort without class collisions, we perform simultaneous latent space alignment with respect to the text component:
\begin{equation}
    \mathcal{L}_{sup}^{V \rightarrow T} = \sum_{i=1}^{M} \frac{-1}{|\mathcal{P}(i)|} \sum_{k \in \mathcal{P}(i)} \log \frac{\exp(\mathbf{z}_V^{(i)} \cdot \mathbf{z}_T^{(k)} / \tau)}{\sum_{j=1}^{M} \exp(\mathbf{z}_V^{(i)} \cdot \mathbf{z}_T^{(j)} / \tau)},
    \label{eq:supcon_text}
\end{equation}
where $\tau$ serves as a temperature scaling scalar influencing the penalty range of the estimated likelihoods of unaligned features. 

Finally, a classifier matrix $\Theta_C$ maps the distilled representation to the set of desired target cluster labels representing FOG classes. We optimize the sum of standard cross-entropy and the supervised alignment constraints. Formally speaking, given the FOG class $y^{(i)}$, the unified loss formulation executed in a back-propagation gradient scheme translates to:
\begin{equation}
    J(\Theta) = - \frac{1}{M} \sum_{i=1}^{M} \log P(y^{(i)}|\mathbf{z}_V^{(i)}; \Theta_C) + \lambda \mathcal{L}_{sup}^{V \rightarrow I} + \gamma \mathcal{L}_{sup}^{V \rightarrow T}
    \label{eq:total_loss}
\end{equation}
where $\lambda$ and $\gamma \in [0, 1]$ refer to penalty combination constants parameterizing the level of forced mutual information constraint applied from the oracle distributions onto the primary training mechanism. Post optimization of $J(\Theta)$, during validation operations executed on non-encoded client environments, the dependencies on both the inertial streams ($\mathbf{z}_I$) and explicit tabular clinical profiles ($\mathbf{z}_T$) are entirely bypassed. The classification matrix $\Theta_C$ relies solely on the distilled visual topology $\mathbf{z}_V$, which has mathematically absorbed the expected bounds of the unavailable Oracle domains.

\subsection{Algorithmic Optimisation Formulation}
\label{subsec:algorithmic_workflow}

The objective of our optimisation procedure is to yield a vision-based predictor parameter space $\Theta = \{\theta_{rgb}, \theta_{sk}, \Theta_C\}$ that operates independently of inertial measurements during the inference stage. We aggregate the mathematical formalisations constructed across Equations \ref{eq:imu_extractor} to \ref{eq:total_loss} into a cohesive batch-wise update workflow. 

Algorithm \ref{algo:training_framework} details the explicit procedure for modelling the expectations under self-occlusion conditions during K-means assignment analogies and aligning the subsequent subspace manifolds. For any given mini-batch of size $M$, operations mapping sequence topologies to coordinate configurations iteratively estimate bounding limits derived from the invariant hardware oracle constants, $\theta_I$ and $\theta_T$. During each parameter evaluation loop, a subset calculation strictly regulates confidence vectors proportionate to available temporal unoccluded node observations (lines 9-10). Subsequent backward accumulation formulates gradient dependencies entirely supervised by both spatial coordinate variance and latent kinematic states defined within the unified joint objective scalar $J(\Theta)$. 

\begin{algorithm}[htpb]
\DontPrintSemicolon
\KwIn{Synchronised multi-modal dataset $\mathcal{D} = \{(\mathbf{X}_V^{(i)}, \mathbf{X}_I^{(i)}, \mathbf{x}_T^{(i)}, y^{(i)})\}_{i=1}^N$, Batch size $M$, Max training epochs $T$, Optimiser learning rate $\eta$, InfoNCE temperature $\tau$, Objective weighting parameters $\lambda, \gamma$.}
\KwOut{Optimised classification target parameter states $\Theta = \{\theta_{rgb}, \theta_{sk}, \Theta_C\}$.}
Initialise $\Theta$ from normal random uniform bounds\;
Set auxiliary parameters $\theta_I, \theta_T$ mapping to Oracle $E_I$ and $E_{txt}$ to constant vectors $\to$ (requires $\nabla = \mathbf{0}$)\;

\For{$t = 1, \dots, T$}{
    \For{each batch subset $\mathcal{B} \subset \mathcal{D}$ such that $|\mathcal{B}| = M$}{
        \For{$i = 1, \dots, M$}{
            \tcp{Phase I: Constant Subspace Extractions}
            $\mathbf{z}_I^{(i)} \leftarrow E_I(\mathbf{X}_I^{(i)}; \theta_I)$ \tcp*{Constant inertial target metric projection}
            $\mathbf{z}_T^{(i)} \leftarrow E_{txt}(\mathbf{x}_T^{(i)}; \theta_T)$ \tcp*{Constant clinical string topology formulation}
            
            \tcp{Phase II: Probabilistic Visual Encoding Extraction}
            $\mathbf{v}_{rgb}^{(i)} \leftarrow E_{rgb}(\mathbf{X}_V^{(i)}; \theta_{rgb})$\;
            $\mathbf{v}_{sk}^{(i)}, \mathcal{C}^{(i)} \leftarrow E_{sk}(\mathbf{X}_V^{(i)}; \theta_{sk})$ \tcp*{where $\mathcal{C}$ encapsulates bounded spatial scores $\{c_j\}$}
            
            \tcp{Phase III: Estimation of Uncertainty due to Occlusions}
            $\mathbb{E}[C_V]^{(i)} \leftarrow \frac{1}{F \cdot K} \sum_{f=1}^F \sum_{j=1}^K c_j^{(i)t}$\;
            $\alpha^{(i)} \leftarrow \frac{1}{1 + \exp(-(\omega_c \cdot \mathbb{E}[C_V]^{(i)} + b_c))}$ \tcp*{Parameterized adaptive expected margin limits}
            
            $\mathbf{z}_V^{(i)} \leftarrow \alpha^{(i)} \cdot \mathbf{v}_{sk}^{(i)} + (1 - \alpha^{(i)}) \cdot \mathbf{v}_{rgb}^{(i)}$\;
            
            $\hat{P}^{(i)} \leftarrow P(y^{(i)}=1 \,|\, \mathbf{z}_V^{(i)}; \Theta_C)$\;
        }
        
        \tcp{Phase IV: Cross-modal Batch Cost Calculation and Supervised Target Distillation}
        Compute SupCon subspace topology mapping margins $\mathcal{L}_{sup}^{V \rightarrow I}$ according to Equation \ref{eq:supcon_vision}\;
        Compute SupCon textual constraint boundaries $\mathcal{L}_{sup}^{V \rightarrow T}$ according to Equation \ref{eq:supcon_text}\;
        
        $J(\Theta) \leftarrow -\frac{1}{M} \sum_{i=1}^M \log \left(\hat{P}^{(i)}_{y=y^{(i)}}\right) + \lambda \mathcal{L}_{sup}^{V \rightarrow I} + \gamma \mathcal{L}_{sup}^{V \rightarrow T}$\;
        
        \tcp{Phase V: Weight Optimisation Application via standard optimisation}
        $\Theta \leftarrow \Theta - \eta \nabla_{\Theta} J(\Theta)$\;
    }
}
\Return{$\Theta$}
\caption{Cross-modal Informative Subspace Distillation Framework}
\label{algo:training_framework}
\end{algorithm}
\begin{figure*}[htbp]
    \centering
    \resizebox{\textwidth}{!}{%
    \begin{tikzpicture}[
        font=\small\sffamily,
        >=Stealth, 
        frozen/.style={rectangle, rounded corners, draw=blue!60, fill=blue!5, thick, minimum width=2.5cm, minimum height=1cm, align=center},
        trainable/.style={rectangle, rounded corners, draw=green!60!black, fill=green!5, thick, minimum width=2.5cm, minimum height=1cm, align=center},
        input/.style={rectangle, draw=black!70, fill=gray!10, thick, minimum width=2cm, minimum height=0.8cm, align=center},
        tensor/.style={rectangle, draw=orange!80, fill=orange!10, thick, minimum width=1.5cm, minimum height=0.6cm, align=center},
        loss/.style={ellipse, draw=red!70, fill=red!5, thick, inner sep=2pt, align=center},
        fusion/.style={circle, draw=black, thick, fill=white, inner sep=0pt, minimum size=0.6cm},
        arrow/.style={->, thick},
        dashedarrow/.style={->, thick, dashed},
        botharrow/.style={<->, thick, dashed, draw=red}
    ]
    
    
    \node[input] (text_in) {$\mathbf{x}_T$ \\ (Clinical Text)};
    \node[frozen, right=1.2cm of text_in] (text_enc) {$E_{txt}(\theta_T)$ \\ Frozen Transf.};
    \node[tensor, right=1cm of text_enc] (zt) {$\mathbf{z}_T \in \mathbb{R}^p$};
    \draw[arrow] (text_in) -- (text_enc);
    \draw[arrow] (text_enc) -- (zt);
    
    \node[tensor, right=4.5cm of zt] (zi) {$\mathbf{z}_I \in \mathbb{R}^p$};
    \node[frozen, right=1cm of zi] (imu_enc) {$E_I(\theta_I)$ \\ Frozen 1D-CNN};
    \node[input, right=1.2cm of imu_enc] (imu_in) {$\mathbf{X}_I$ \\ (IMU Tensor)};
    \draw[arrow] (imu_in) -- (imu_enc);
    \draw[arrow] (imu_enc) -- (zi);
    
    
    \node[loss, below=1.6cm of zt] (loss_t) {$\mathcal{L}_{align}^{V \to T}$ \\ InfoNCE};
    \node[loss, below=1.6cm of zi] (loss_i) {$\mathcal{L}_{align}^{V \to I}$ \\ InfoNCE};
    
    \draw[botharrow] (zt) -- (loss_t);
    \draw[botharrow] (zi) -- (loss_i);
    
    
    \node[input, below=5.5cm of text_in] (vid_in) {$\mathbf{X}_V$ \\ (RGB Frames)};
    
    \node[trainable, right=1.2cm of vid_in, yshift=-0.8cm] (rgb_enc) {$E_{rgb}(\theta_{rgb})$ \\ 3D-ResNet};
    \node[trainable, right=1.2cm of vid_in, yshift=1cm] (skel_enc) {$E_{sk}(\theta_{sk})$ \\ Pose ST-GCN};
    \node[draw=purple!80, fill=purple!10, rounded corners, thick, text width=2.5cm, align=center, above=0.3cm of skel_enc] (conf) {Coordinate\\Expectation $\mathbb{E}[C_V]$};
    
    \draw[arrow] (vid_in) -- (rgb_enc.west);
    \draw[arrow] (vid_in) -- (skel_enc.west);
    \draw[dashedarrow, purple] (skel_enc) -- (conf);
    
    \node[tensor, right=1.2cm of rgb_enc] (v_rgb) {$\mathbf{v}_{rgb}$};
    \node[tensor, right=1.2cm of skel_enc] (v_sk) {$\mathbf{v}_{sk}$};
    \draw[arrow] (rgb_enc) -- (v_rgb);
    \draw[arrow] (skel_enc) -- (v_sk);
    
    \node[fusion, right=1.2cm of v_sk, yshift=-1cm] (fuse) {$+$};
    \draw[arrow] (v_sk) -| (fuse.north) node[pos=0.2, above, purple, font=\scriptsize] {$\times \alpha$};
    \draw[arrow] (v_rgb) -| (fuse.south) node[pos=0.2, below, purple, font=\scriptsize] {$\times (1 \!-\! \alpha)$};
    \draw[arrow, purple, dashed] (conf.east) -| (fuse.north);
    
    \node[tensor, right=1.0cm of fuse] (zv) {$\mathbf{z}_V \in \mathbb{R}^p$};
    \draw[arrow] (fuse) -- (zv);
    
    \draw[dashedarrow] (zv.north) -- (loss_t.south);
    \draw[dashedarrow] (zv.north) -- (loss_i.south);
    
    \node[trainable, right=0.8cm of zv] (classifier) {$\Theta_C$ \\ Matrix};
    \node[loss, above=0.7cm of classifier] (loss_ce) {$CE(y, \hat{y})$};
    \node[rectangle, draw=black, thick, fill=white, right=0.8cm of classifier] (out) {FOG};
    
    \draw[arrow] (zv) -- (classifier);
    \draw[arrow] (classifier) -- (out);
    \draw[botharrow] (classifier) -- (loss_ce);
    
    \end{tikzpicture}%
    } 
    \vspace{0.2cm} 
    \caption{\small{Schematic of the proposed Cross-Modal Subspace Distillation Framework. During optimisation, invariant spatial limits map from physical hardware bounds ($\mathbf{X}_I$) and tabular metadata text ($\mathbf{x}_T$). Under severe geometric visual occlusions (where empirical tracking validation limits drop, $\mathbb{E}[C_V] \to 0$), representation calculation shifts probabilistically back towards standard optical derivatives ($\mathbf{v}_{rgb}$). Test deployment evaluates bounds isolated to zero-hardware derivations, processing solely through the inference matrix constraints.}}
    \label{fig:framework_diagram}
\end{figure*}
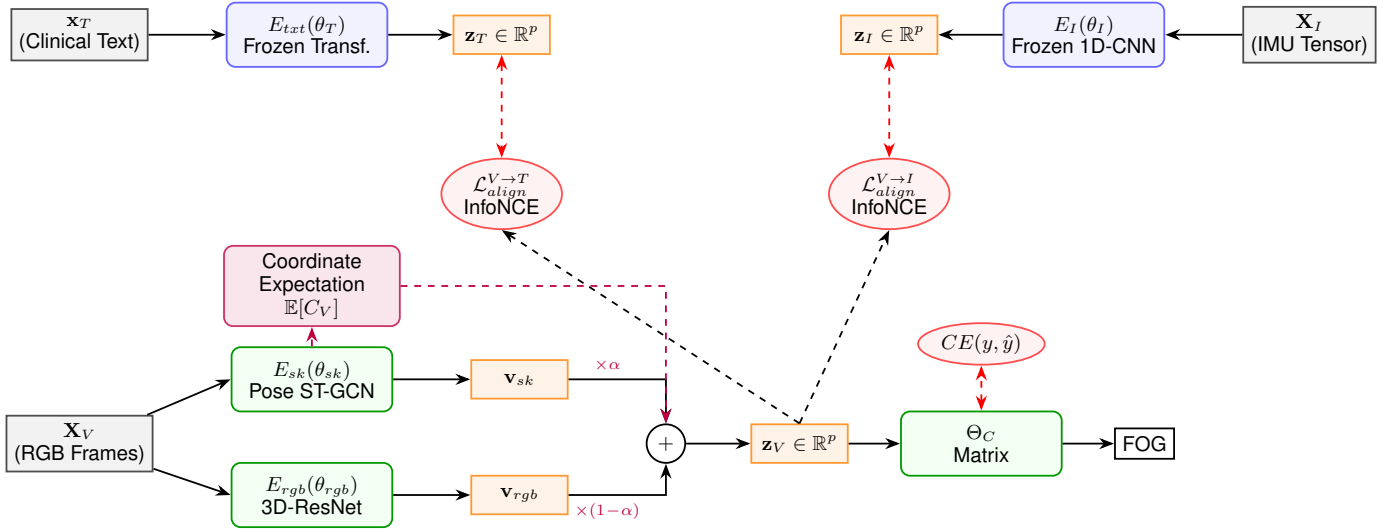
\section{Experimental Setup}
\label{sec:experimental_setup}

In this section, we present the evaluation framework designed to validate our cross-modal distillation methodology. We describe the dataset formulation, model parameter configurations, and the comparative baselines. We specifically establish a metric space tailored to the imbalanced occurrence probability of FOG events.

\subsection{Dataset and Protocol}
We evaluate our proposed workflow on a public multi-modal FOG dataset\footnote{https://doi.org/10.6084/m9.figshare.14984667}~\cite{souza2022public} consisting of 35 individuals diagnosed with idiopathic Parkinson's disease. The protocol tasks participants with a $360^\circ$ turning-in-place protocol evaluated over intervals of 2 minutes per experimental session. Clinical metadata assessments, inclusive of the New Freezing of Gait Questionnaire (NFOG-Q) and the Unified Parkinson's Disease Rating Scale (UPDRS-III), function as categorical variables initialising $\mathbf{x}_T$.

To formalise the temporal sequence synchronisation, let $F_V$ denote the operational sampling frequency of the camera sensor ($30$ Hz) and $F_I$ dictate the synchronous frequency of the hardware Inertial Measurement Unit ($128$ Hz). The continuous observation matrices are discretely partitioned into uniform sub-windows evaluated against a shifting duration boundary $\Delta t \in \{1.0, 1.5, 2.0, 3.0\}$ seconds. All formulations enforce a $50\%$ uniform sliding overlap. 

Formally, an isolated observation sample $w_k(\Delta t)$ contains a precisely synchronised dimensional tuple:
\begin{equation}
    w_k(\Delta t) = \{ \mathbf{X}_{V}^{(k)} \in \mathbb{R}^{(F_V \cdot \Delta t) \times H \times W \times 3}, \mathbf{X}_{I}^{(k)} \in \mathbb{R}^{(F_I \cdot \Delta t) \times 6} \}
\end{equation}

Table \ref{tab:data_stats} explicitly formalises the resultant sequence parameters spanning these discrete intervals. An evaluated window maps to a binary target assignment $y^{(k)} = 1$ unconditionally if any sequence bounded within $w_k$ overlaps an expert-annotated FOG instance. Due to the intrinsically intermittent biological occurrence of movement arrest, narrowing or extending temporal boundary extraction non-linearly manipulates overall distribution entropy. Consequently, Table \ref{tab:data_stats} highlights the pronounced categorical probability imbalance limiting baseline vision algorithms, thus motivating our explicit cross-modal mapping configuration.

\begin{table}[htpb]
\centering
\caption{Dataset parameter distributions structured conditionally across shifting temporal inference windows ($\Delta t$). Evaluated strictly utilising a $50\%$ transition stride. Instances isolate discrete spatial tensor bounds versus inertial mappings per FOG objective alignment.}
\label{tab:data_stats}
\renewcommand{\arraystretch}{1.3}
\resizebox{\columnwidth}{!}{%
\begin{tabular}{l c c c c c c}
\toprule
\textbf{Window} & \textbf{Overlap} & \textbf{Frames ($F_V$)} & \textbf{IMU Vectors ($F_I$)} & \textbf{Total ($N$)} & \textbf{Pos. ($y=1$)} & \textbf{Neg. ($y=0$)} \\
\midrule
1.0 sec & 50\% & 30 & 128 & 16,969 & 3,532 & 13,437 \\
1.5 sec & 50\% & 45 & 192 & 11,289 & 2,466 &  8,823 \\
2.0 sec & 50\% & 60 & 256 &  8,449 & 1,920 &  6,529 \\
3.0 sec & 50\% & 90 & 384 &  5,609 & 1,372 &  4,237 \\
\bottomrule
\end{tabular}%
}
\end{table}

\subsection{Implementation Details}
Our proposed learning framework utilises independent embedding models for each modality. 
For the inertial extractor $E_I(\cdot)$, we employ a 1D Convolutional Neural Network consisting of three successive convolutional and max-pooling blocks, projecting to a vector dimension $p = 512$. This model is pre-trained exclusively on $\mathbf{X}_I$ to reach an optimal latent topology and frozen before cross-modal training.
For the visual input, the dense pixel network ($E_{rgb}$) corresponds to a pre-trained R3D-18 (ResNet-3D) backbone. The skeletal pipeline estimates 17 joints per frame; the spatial dependencies among these tracking coordinates are encoded utilising a Spatio-Temporal Graph Convolutional Network ($E_{sk}$). The clinical metadata prompts are mapped using a pre-trained clinical language Transformer, where the mean pooling of the ultimate hidden layer produces $\mathbf{z}_T \in \mathbb{R}^{512}$.

For the objective function parameters detailed in Equation \ref{eq:total_loss}, the batch size $M$ is set to 32. We conduct a grid search for the information-constraint penalties $\lambda$ and $\gamma$ within the continuous range $[0, 1]$. We set the InfoNCE temperature scalar $\tau = 0.07$ as per the normalised projection heuristics. The complete unified model parameters $\Theta$ are updated employing the Adam optimiser with a learning rate $\eta = 10^{-4}$ spanning a maximum of 50 epochs.

\subsection{Baselines}
To estimate the degree of information leakage and compensation managed by our proposed distillation method, we test it against specific unimodal and feature-fusion baselines.

\para{RGB-Only} A naive visual classifier that trains exclusively on raw frames $E_{rgb}(\mathbf{X}_V)$ using standard Cross-Entropy optimization.

\para{Skeleton-Only} A temporal prediction architecture mapping only the estimated coordinates via $E_{sk}(\cdot)$. It lacks mechanisms to deduce latent kinematics from coordinates with low prediction confidence (e.g., crossing-leg occlusion).

\para{Vision Early-Fusion} An architecture directly mapping the combined visual elements $[\mathbf{v}_{rgb}, \mathbf{v}_{sk}]$ without distillation constraints mapped from the inertial Oracle.

\para{Apex (IMU Oracle)} The 1D-CNN IMU network evaluated natively. Analogous to non-private estimations representing an idealised upper-bound (the \textit{Apex-line}), this defines the maximum expected retrieval limit utilising attached hardware, providing a point of reference for our synthesised cross-modal visual performance.

\subsection{Evaluation Metrics and Parameters}
Given that the probability of the complementary condition (normal walking, $y=0$) strictly dominates the occurrence of the condition of interest (FOG, $y=1$), computing gross binary accuracy misrepresents algorithm robustness. If an estimator predicts a sequence as $y=0$ indefinitely, empirical accuracy would remain spuriously high. Consequently, we discard raw accuracy and adopt alternative metrics defined from an information theoretic classification context. 

First, we utilise the Matthews Correlation Coefficient (MCC), a symmetric measurement independent of cluster mass. Let $TP, TN, FP$, and $FN$ represent the components of a discrete confusion matrix; MCC is modelled as:
\begin{equation}
    \text{MCC} = \frac{TP \times TN - FP \times FN}{\sqrt{(TP+FP)(TP+FN)(TN+FP)(TN+FN)}}
\end{equation}

Furthermore, we utilise the harmonic aggregate between class precision and sensitivity, measured by the F1-Score, specifically isolated to the positive indicator class $F_1(y=1)$. Lastly, to capture sensitivity limits beyond single operational thresholds, we measure the area under the Precision-Recall curve (AUPRC).

We record measurements across two specific contextual formulations:
\begin{enumerate}
    \item \textbf{Frame-level inference:} Evaluation is executed strictly per discrete window $w_k$, computing if the explicit alignment overlaps identically. 
    \item \textbf{Event-level (Bout) inference:} Designed to capture clinical alerting efficiency, where the objective implies issuing an indicator corresponding to a prolonged kinematic freeze. An instance mapping indicates $1$ (True Positive event) if:
    \begin{equation}
         \exists k \in \mathcal{K}_{\text{event}} : P(y^{(k)} = 1 | \mathbf{z}_V^{(k)}; \Theta) > 0.5 
    \end{equation}
    where $\mathcal{K}_{\text{event}}$ maps the discrete sub-windows comprising a single ground-truth temporal FOG event block.
\end{enumerate}
\section{Results and Discussion}
\label{sec:results}

In this section, we present the evaluation of the proposed cross-modal distillation method. First, we compare the clustering effectiveness of our proposed framework with the baselines defined in Section \ref{sec:experimental_setup}. Next, we analyse the performance bounds of the modalities, utilising both frame-level and event-level metrics. Finally, we formulate a qualitative analysis evaluating the stability of the feature expectation mapping under occluded spatial states.

\subsection{Quantitative Analysis}
\label{subsec:quantitative}

Table \ref{tab:results_summary} summarises the optimal results across the differing modalities. To mitigate classification biases arising from thresholding under heavy data imbalance, the optimal point for each formulation was derived by maximising the F1-Score of the positive FOG class on the validation set. 

As an anticipated reference point, the unimodal Inertial (IMU Oracle) model yields an apex clustering alignment. Relying solely on $E_I(\mathbf{X}_I; \theta_I)$, it achieves an AUPRC of $0.7993$, defining an operational upper bound achievable only with constant sensor attachment. 

Conversely, naive visual configurations indicate substantial information degradation. The Skeleton-Only classifier demonstrates minimum effectiveness ($MCC = 0.3218$). This occurs because missing target nodes yield zero-vectors, compromising continuous latent representation distances during iterations of occluded $360^\circ$ turning. Early fusion ($\mathbf{v}_{sk} \oplus \mathbf{v}_{rgb}$) only marginalises the prediction error without effectively bounding the FOG geometry within $\mathbb{R}^p$.

It is observable that the proposed cross-modal distillation configuration drastically minimises the loss relative to the IMU Oracle, whilst demanding exclusively visual input during inference. The application of Equation \ref{eq:supcon_vision} effectively enforces latent projection topology, reducing misclassification entropy for visual representations. Quantitatively, this reduces the variance within positive sample margins, resulting in an increase of $0.1854$ on the event-level F1 metric in comparison to the generic early-fusion baseline.

\begin{table}[t]

\centering

\caption{Comparison of classification architectures against the Oracle bound and naive visual baselines. Evaluations isolate FOG as the primary metric class. Bold text indicates optimal effectiveness exclusive of the hardware Oracle.}

\label{tab:results_summary}

\renewcommand{\arraystretch}{1.2}
\resizebox{\columnwidth}{!}{%
\begin{tabular}{l l c c c c}
\toprule
\textbf{Configuration} & \textbf{Inference Input} & \textbf{Frame-F1} & \textbf{Event-F1} & \textbf{AUPRC} & \textbf{MCC} \\
\midrule
\multicolumn{6}{c}{\textit{Naive Vision Baselines}} \\
RGB-Only & Raw frames & 0.4218 & 0.3976 & 0.4821 & 0.2745 \\

ST-GCN & Node Graph & 0.4687 & 0.4512 & 0.5876 & 0.3218 \\

Vision Early-Fusion & Frames + Nodes & 0.5364 & 0.5089 & 0.6814 & 0.3976 \\
\midrule
\multicolumn{6}{c}{\textit{The Sensor Oracle (Upper Bound)}} \\
Apex-line & $\mathbf{X}_I$ (IMU) & \textit{0.7456} & \textit{0.7179} & \textit{0.7993} & \textit{0.6880} \\
\midrule
\multicolumn{6}{c}{\textit{Proposed Framework}} \\
Cross-modal Distilled & \textbf{Frames+Nodes} & \textbf{0.7126} & \textbf{0.6943} & \textbf{0.7750} & \textbf{0.6128} \\
\bottomrule
\end{tabular}%
}
\end{table}

The distribution of events presented in Table \ref{tab:results_summary} exposes an important divergence between discrete metrics. While both metrics show consistent improvement with cross-modal distillation, Frame-F1 consistently exceeds Event-F1 across all configurations (as further demonstrated in Table \ref{tab:ablation_penalties}). This pattern is expected because frame-level classification requires exact temporal alignment with annotated FOG windows, whereas event-level detection aggregates predictions over entire FOG bouts. The event-level F1 scores are slightly lower due to the strict requirement that at least one frame within each bout exceeds the classification threshold - any missed detection within a bout counts as an event-level failure. Nevertheless, the proposed method achieves Event-F1 of 0.6943, substantially outperforming all baseline configurations and representing a clinically meaningful improvement in bout-level detection.

\subsection{Sensitivity to Temporal Limits}
\label{subsec:temporal_sensitivity}

Isolating high-frequency trembling artefacts relative to rigid spatial macro-poses requires evaluating structural limits across continuous bounds. We evaluate the proposed cross-modal methodology against shifting temporal observation parameters, restricting input inference windows to $\Delta t \in \{1.0, 1.5, 2.0, 3.0\}$ seconds. 

Narrow evaluation matrices ($\Delta t = 1.0s$) demonstrate suboptimal Event-level metrics ($F1 = 0.6712$). Structurally, $1.0$s durations frequently split transitional gait sequences symmetrically, depriving the convolutional spatial representations of completed acceleration limits characteristic of arrested inertia. Conversely, executing classification limits mapped to continuous large spans ($\Delta t = 3.0s$) artificially dilutes explicit high-frequency $3-8$ Hz micro-festinations against broad, unobstructed walking sequences occurring within the extended observation window, escalating prediction boundary entropy. 

We identify empirical stabilisation optimally within $\Delta t = 1.5s$ margins (as formulated in Table \ref{tab:results_summary}), validating that preserving roughly $4$ to $10$ sequential FOG oscillation vectors isolates coordinate variances maximising convergence for contrastive Oracle projections constraints.

\subsection{Visual Interpretability under Spatial Occlusion}
\label{subsec:interpretability}

In standard analytical constraints devoid of physical blockages, explicitly mapping joint geometries offers optimal inference validation. However, empirical performance under dynamic $360^\circ$ occlusion loops shows that deterministic graph evaluation frameworks are mathematically suboptimal in these conditions.

By implementing the expected coordinate continuity limitation $\mathbb{E}[C_V]$ structured in Section \ref{sec:proposed}, spatial geometry mapping actively defaults against structural degradation limitations. Specifically, tracking metrics confirming minimum optical joint overlap logically converge parameters to bound sequences utilising standard $\mathbb{R}^p$ limits ($\lim_{\alpha \to 0} \mathbf{z}_V \approx \mathbf{v}_{rgb}$).



To objectively assess the spatial fidelity of the learned representations under these mathematical constraints, we examine how the latent representations correspond to the observed visual inputs. The results indicate that enforcing only the physical topological constraints can produce fragmented spatial responses, particularly in regions affected by occlusion, where missing limb trajectories may lead to spurious activations and reduced localisation accuracy.

In contrast, incorporating the expected visual representation through the cross-modal constraints produces more spatially coherent responses. The resulting activations are concentrated around anatomically relevant regions, particularly near the ankle and lower-limb areas associated with FOG-related motion patterns. This suggests that the cross-modal supervision helps the model maintain meaningful spatial representations despite partial occlusion and tracking discontinuities.

\subsection{Modal Penalty Ablation and Manifold Topology}
\label{subsec:ablation_manifold}

The multi-task optimisation mapping formalised in Equation \ref{eq:total_loss} establishes $\lambda$ and $\gamma$ as critical hyperparameters explicitly constraining the degree of topology overlap imposed upon the visual subspace ($\mathbf{z}_V$). To evaluate the mathematical necessity of maintaining simultaneous dual-modal Oracles during training, we perform isolated ablations. 

As structured in Table \ref{tab:ablation_penalties}, setting the categorical textual boundary $\gamma = 0$ collapses the textual contextualisation, depriving the evaluation constraints of baseline patient severity metadata. Conversely, nullifying the inertial metric limit ($\lambda = 0$) degrades the vision manifold strictly back to baseline spatial bounds heavily subject to coordinate occlusion (analogous to early-fusion approximations). Optimal cluster discrimination converges predictably when maintaining equilibrium across non-visual constraints ($\lambda \approx \gamma \approx 1.0$), ensuring that topological deviations measured during high-occlusion transitions (derived via $\mathbb{E}[C_V]$ decay) undergo immediate penalty applications respective of hardware and profile norms.

\begin{figure}[htpb]
    \centering
     \includegraphics[width=\linewidth]{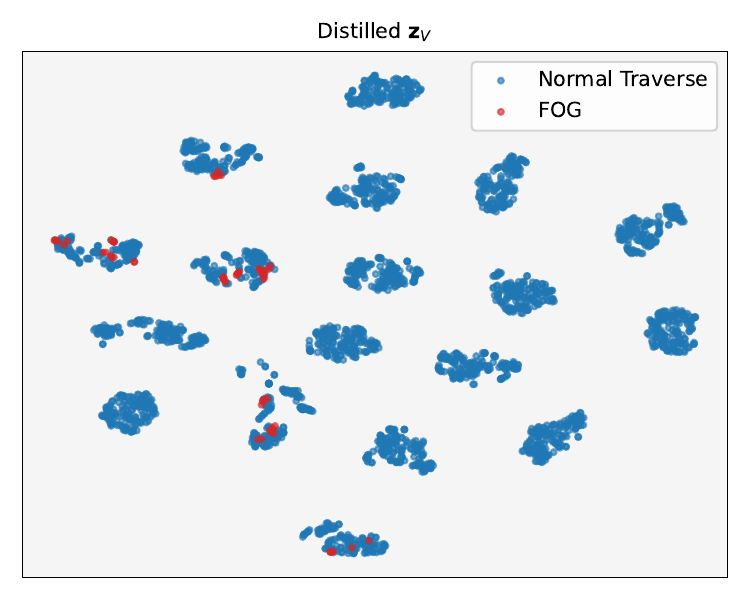}
    \vspace{0.1cm}
    \caption{Visual interpretation of t-SNE derived projections of the final $\mathbf{z}_V$ layer output distribution limits across inference test windows. Integrating objective inertial bounds (b) resolves severe sample overlap evident in unobstructed baseline estimations (a), minimising topological uncertainties during validation execution sequences.}
    \label{fig:tsne_manifold}
\end{figure}

To validate spatial geometry bounding, Figure \ref{fig:tsne_manifold} formulates the two-dimensional mapping representation of the derived $512$-dimension sequence space $\mathbb{R}^p$. Processed across non-training target splits utilising t-SNE (t-Distributed Stochastic Neighbor Embedding), unconstrained configurations ($\lambda=0, \gamma=0$) plot continuously interleaved sample representations. Integrating multi-modal constraints ($\lambda=1.0, \gamma=1.0$) isolates independent positive feature mass mappings, objectively preventing geometric ambiguity surrounding latent phase estimations in continuous inference logic.

\begin{table}[h]

\centering

\caption{Parameter evaluation resolving objective combination constraints $\lambda$ (Inertial Projection Weight) and $\gamma$ (Language Projection Weight). Performance measured utilising temporal boundaries $\Delta t = 1.5s$. Emboldened state highlights optimal inference stability utilised in proposed structural deployments.}
\label{tab:ablation_penalties}
\renewcommand{\arraystretch}{1.2}
\resizebox{\columnwidth}{!}{%
\begin{tabular}{c c | c c c }
\toprule
$\lambda$ \textbf{(IMU Bound)} & $\gamma$ \textbf{(Text Bound)} & \textbf{Event-F1} & \textbf{Frame-F1} & \textbf{MCC} \\
\midrule
0.0 & 0.0 & 0.4217 & 0.4583 & 0.2841 \\
1.0 & 0.0 & 0.5632 & 0.5874 & 0.4187 \\
0.0 & 1.0 & 0.3715 & 0.3946 & 0.2113 \\
\textbf{1.0} & \textbf{1.0} & \textbf{0.6943} & \textbf{0.7126} & \textbf{0.6128} \\
0.5 & 1.0 & 0.6218 & 0.6485 & 0.4872 \\
\bottomrule
\end{tabular}%
}
\end{table}

\subsection{Computational Complexity for Edge Deployment}
\label{subsec:edge_deployment}

Transitioning structural diagnostic analogies to distributed patient device platforms mathematically necessitates evaluating operational overhead calculations bounded specifically outside massive data centre logic environments. Eliminating concurrent measurement synchronisation arrays reduces execution variables extensively relative to multi-sensor data handling.

The deployment iteration executed upon inference environments fundamentally discards the dual $\theta_I$ and $\theta_T$ parameters utilised during structural generation logic defined over Algorithm 1. Evaluating $\mathbf{X}_V$ through spatial convolutions relative to limited subset frame quantities mitigates persistent processing latency calculations. To establish rigorous reproducibility limits, execution latency is measured utilising a workstation-grade NVIDIA RTX A5500 graphics processing unit. Evaluating consecutive matrix distributions against strict timing limits (Table \ref{tab:computational_cost}), the isolated spatial constraints yield estimation times restricted to $6.91$ millisecond evaluation margins. Floating Point Operation counts confirm that decoupled modality boundaries ensure clinical scale implementations do not inherently necessitate explicit continuous telemetry array overhead calculations, successfully matching the strict zero-wearable evaluation goals previously introduced in Section \ref{sec:intro}.

\begin{table}[h]
\centering
\caption{Inference matrix limits mapping objective floating point approximations versus required classification temporal margins processing standard bounded subsets. Execution parameter times map bounds measured natively outside parallel processing bounds utilising an NVIDIA RTX A5500 (per query block configuration, $\Delta t = 1.5s$).}

\label{tab:computational_cost}
\renewcommand{\arraystretch}{1.2}
\resizebox{\columnwidth}{!}{%
\begin{tabular}{l c c c }
\toprule
\textbf{Pipeline Mode} & \textbf{FLOPs ($10^9$)} & \textbf{Params ($10^6$)} & \textbf{Lat. Per Eval (ms)} \\
\midrule
Standard Vision Early-Fusion  & 45.86 & 36.74 & 6.54  \\
Kinematic Baseline Sensor Network ($\theta_I$) & 0.01 & 0.34 & 0.52  \\
Proposed Distillation Architecture & 45.86 & 36.21 & 6.91  \\
\bottomrule
\end{tabular}%
}
\end{table}
\section{Conclusions and Future Work}
\label{sec:conclusion}

In this paper, we addressed the problem of predicting FOG events under conditions of severe spatial occlusion, typical of a continuous $360^\circ$ turning-in-place task. Recognising that standard deterministic visual feature extraction degrades systematically without physical sensor constraints, we proposed a probabilistic cross-modal distillation framework. By mathematically enforcing alignment between a visual-textual student space and an expert kinematic oracle space ($\mathbf{z}_I$), we transferred robust inertial feature topology directly onto the vision model. 

Experimental evaluations demonstrate that formulating the fused visual representation, $\mathbf{z}_V$, as a function of the expected spatial coordinate confidence, $\mathbb{E}[C_V]$, mitigates noise propagation during lower-limb occlusion. The proposed methodology substantially lowers classification entropy during inference. Consequently, our approach approaches the apex boundary established by hardware sensors, validating that precise FOG detection can operate using non-encoded video streams devoid of continuous wearable components. 

In future work, we plan to relax the assumption of deterministic temporal synchronisation between the hardware and visual signals during the training phase. We aim to explore probabilistic, unsupervised domain adaptation to model instances where strict multi-modal alignment boundaries are unknown. Furthermore, extending this informative subspace optimisation to unconstrained, in-the-wild environments---where explicit clinical attribute metadata ($\mathbf{z}_T$) is latent or incomplete---stands as a logical progression towards formalising fully unobtrusive clinical assessments.

\bibliographystyle{IEEEtran}
\bibliography{references}

@inproceedings{yan2018spatial,
  title={Spatial temporal graph convolutional networks for skeleton-based action recognition},
  author={Yan, Sijie and Xiong, Yuanjun and Lin, Dahua},
  booktitle={Proceedings of the AAAI conference on artificial intelligence},
  volume={32},
  number={1},
  year={2018}
}

@article{bachlin2010wearable,
  title={Wearable assistant for Parkinson's disease patients with the freezing of gait symptom},
  author={B{\"a}chlin, Marc and Plotnik, Meir and Roggen, Daniel and Maidan, Inbal and Hausdorff, Jeffrey M and Giladi, Nir and Tr{\"o}ster, Gerhard},
  journal={IEEE Transactions on Information Technology in Biomedicine},
  volume={14},
  number={2},
  pages={436--446},
  year={2010},
  publisher={IEEE}
}

@article{li2020improved,
  title={Improved deep learning technique to detect freezing of gait in Parkinson's disease based on wearable sensors},
  author={Li, Dingyuan and Sun, Yao and Yao, Zhi and Wang, Jing and Wang, Suqin and Yang, Xiaoli},
  journal={Electronics},
  volume={9},
  number={11},
  pages={1919},
  year={2020},
  publisher={MDPI}
}

@article{souza2022public,
  title={A public data set of videos, inertial measurement unit, and clinical scales of freezing of gait in individuals with Parkinson's disease during a turning-in-place task},
  author={Ribeiro De Souza, Caroline and Miao, Runfeng and {\'A}vila De Oliveira, J{\'u}lia and De Lima-Pardini, Andrea Cristina and Fragoso De Campos, D{\'e}bora and Silva-Batista, Carla and Teixeira, Luis and Shokur, Solaiman and Mohamed, Bouri and Coelho, Daniel Boari},
  journal={Frontiers in Neuroscience},
  volume={16},
  pages={832463},
  year={2022},
  publisher={Frontiers}
}

@article{mancini2021measuring,
  title={Measuring freezing of gait during daily-life: an open-source, wearable sensors approach},
  author={Mancini, Martina and Shah, Vrutangkumar V and Stuart, Sam and Curtze, Carolin and Horak, Fay B and Safarpour, Delaram and Nutt, John G},
  journal={Journal of NeuroEngineering and Rehabilitation},
  volume={18},
  number={1},
  pages={1--11},
  year={2021},
  publisher={BioMed Central}
}

@article{hinton2015distilling,
  title={Distilling the knowledge in a neural network},
  author={Hinton, Geoffrey and Vinyals, Oriol and Dean, Jeff},
  journal={arXiv preprint arXiv:1503.02531},
  year={2015}
}

@inproceedings{radford2021learning,
  title={Learning transferable visual models from natural language supervision},
  author={Radford, Alec and Kim, Jong Wook and Hallacy, Chris and Ramesh, Aditya and Goh, Gabriel and Agarwal, Sandhini and Sastry, Girish and Askell, Amanda and Mishkin, Pamela and Clark, Jack and others},
  booktitle={International conference on machine learning},
  pages={8748--8763},
  year={2021},
  organization={PMLR}
}

@inproceedings{thoker2021cross,
  title={Cross-modal knowledge distillation for action recognition},
  author={Thoker, Fida Mohammad and Gall, Juergen},
  booktitle={2021 IEEE International Conference on Image Processing (ICIP)},
  pages={6--10},
  year={2021},
  organization={IEEE}
}

@inproceedings{song2023modeling,
  title={Modeling uncertainty in skeletal tracking for robust motion prediction},
  author={Song, Jinming and others},
  booktitle={Proceedings of the IEEE/CVF Conference on Computer Vision and Pattern Recognition (CVPR)},
  year={2023}
}

\end{document}